\documentclass[10pt,twocolumn,letterpaper]{article}

\usepackage[pagenumbers]{cvpr} 

\usepackage{etoolbox}
\usepackage{algorithm}
\usepackage{algpseudocode}

\definecolor{cvprblue}{rgb}{0.21,0.49,0.74}
\usepackage[pagebackref,breaklinks,colorlinks,allcolors=cvprblue]{hyperref}

\def\paperID{*****} 
\def\confName{MIV at CVPR}
\def\confYear{2025}

\title{Embedding Shift Dissection on CLIP: Effects of Augmentations on VLM's Representation Learning}

\author{Ashim  Dahal \quad Saydul Akbar Murad \quad Nick Rahimi\\
University of Southern Mississippi\\
Hattiesburg, Mississippi, USA\\
{\tt\small \{ashim.dahal,saydulakbar.murad,nick.rahimi\}@usm.edu}
}

\begin{document}

\twocolumn[{%
\renewcommand\twocolumn[1][]{#1}%
\maketitle
\begin{center}
    \centering
    \captionsetup{type=figure}
    \includegraphics[width=\textwidth]{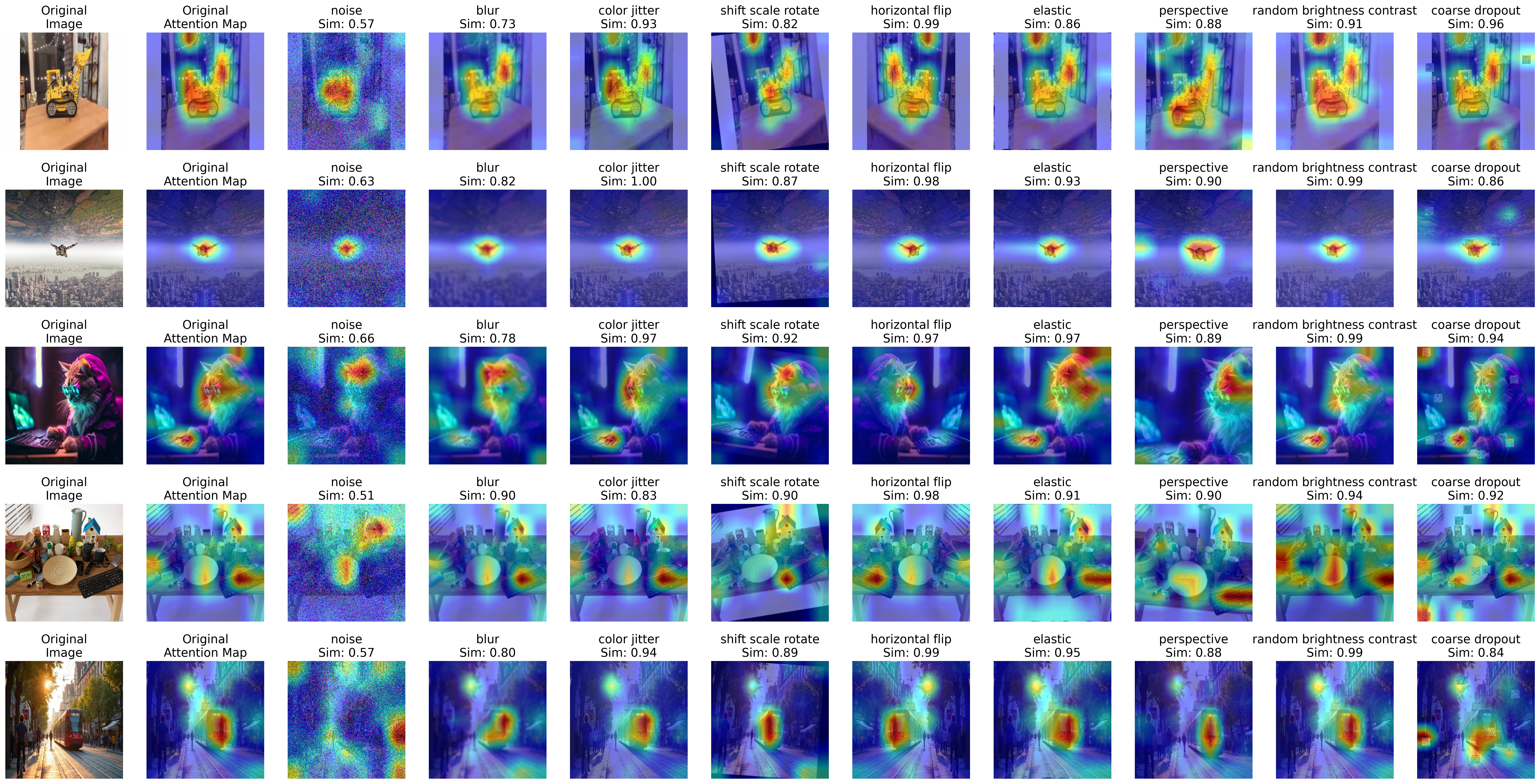}
    \captionof{figure}{Qualitative analysys of final layer of attention map of CLIP for vision augmentation techniques}
    \label{fig:qualitative_results}
\end{center}%
}]

\begin{abstract}
  Understanding the representation shift on Vision Language Models like CLIP under different augmentations provides valuable insights on Mechanistic Interpretability. In this study, we show the shift on CLIP's embeddings on 9 common augmentation techniques: noise, blur, color jitter, scale and rotate, flip, elastic and perspective transforms, random brightness and contrast, and coarse dropout of pixel blocks. We scrutinize the embedding shifts under similarity on attention map, patch, edge, detail preservation, cosine similarity, L2 distance, pairwise distance and dendrogram clusters and provide qualitative analysis on sample images. Our findings suggest certain augmentations like noise, perspective transform and shift scaling have higher degree of drastic impact on embedding shift. This study provides a concrete foundation for future work on VLM's robustness for mechanical interpretation and adversarial data defense. The code implementation for this study can be found on \href{https://github.com/ashimdahal/clip-shift-analysis}{https://github.com/ashimdahal/clip-shift-analysis}.

\end{abstract}

\section{Introduction}
\begin{figure*}
    \centering
    \includegraphics[width=\linewidth]{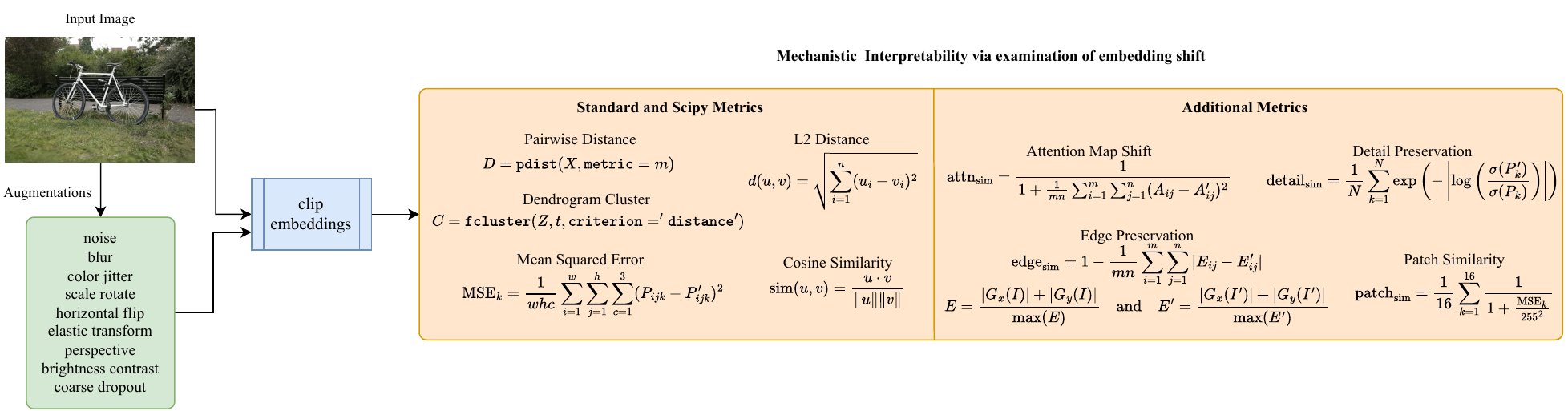}
    \caption{Research methodology and list of qualitative analysis performed}
    \label{fig:methodology}
\end{figure*}
\label{sec:intro}
Vision Language Models (VLM) \cite{vlm} such as Contrastive Language Image Pretraining (CLIP) \cite{clip} have provided a strong generalization of representing images and text in the same latent space. However, studies on their internal representation are scarce. Models like Stable Diffusion \cite{stable_diffusion} use CLIP internally and often perform poorly when standard image augmentations are applied in the image. Our research on Mechanic Interpretability analyses how exactly these representations, their attention, and preservation qualities tend to shift when presented with image augmentations.

Existing works on CLIP's embedding shift understanding focus on understanding the effects of text artifacts and performance evaluation but lack insights into how augmentations under the same image affect the learned representations. The major question we answer is whether augmentation alter the semantic understandings (if they do so then by how much for each augmentation) or if the model shows invariance in its embedding representation. We find the former to be the prominent outcome in most of our 9 augmentation techniques. In summary, we:
\begin{itemize}
    \item Perform systematic analysis of CLIP's response to augmentations
    \item Measure representation drift, attention shift and detail preservation 
    \item Provide qualitative and quantitative analysis like similarity score, patch similarity, edge preservation, detail preservation and dendrogram clusters
    \item Provide concrete insights into how CLIP encode visual transformations and discuss pathway for future research
\end{itemize}

\section{Related Work}\label{sec:lit_review}
\begin{figure}[t]
    \centering
    \includegraphics[width=\linewidth]{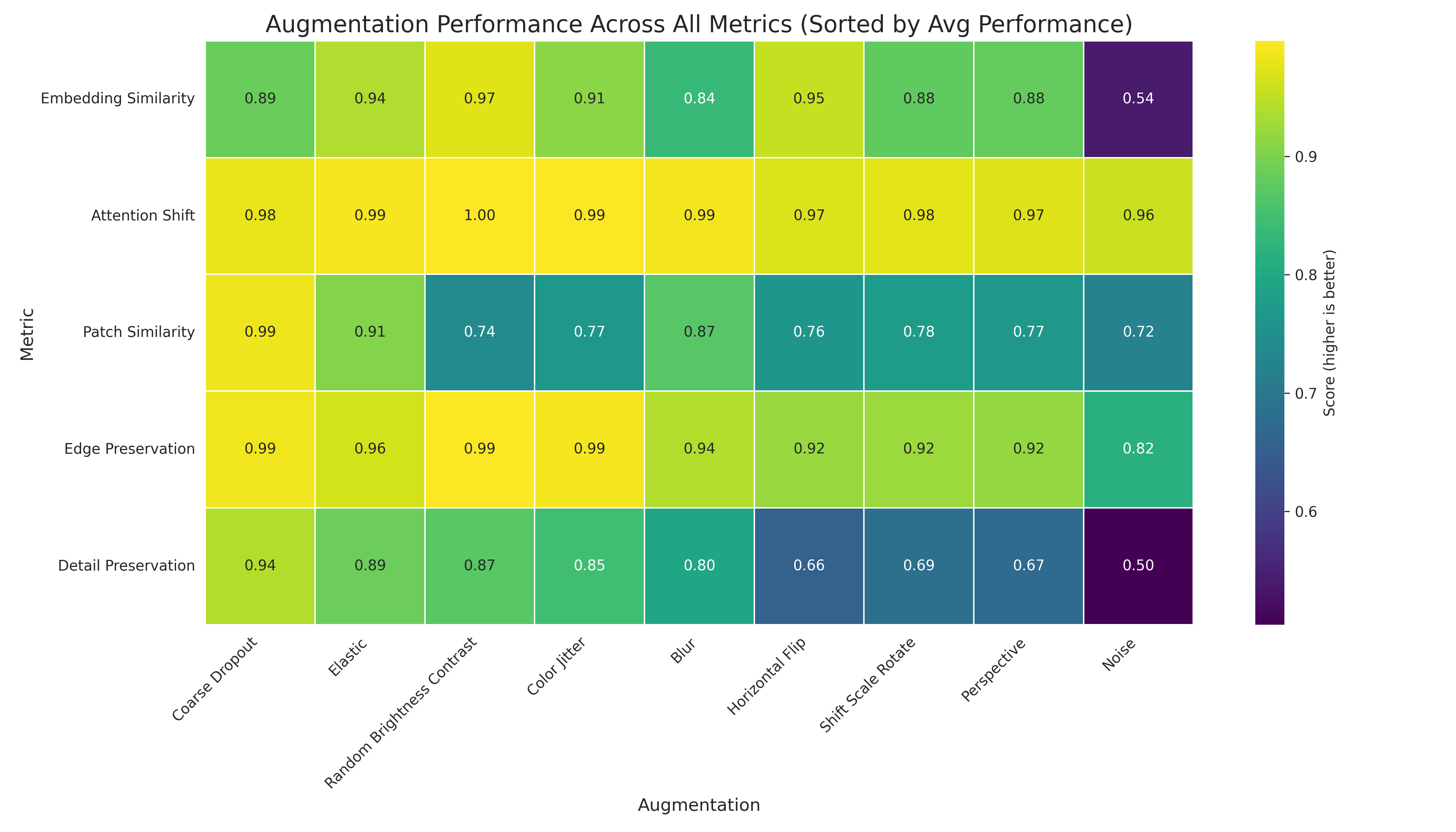}
    \caption{Sorted heatmap of augmentation performance}
    \label{fig:density_sorted}
\end{figure}

Previous research under CLIP-like foundational model's exploration primarily focuses on interpreting the relationship between text and image embeddings \cite{intel}. In those few research studies which focus on the exploration of visual interpretability of CLIP, the effect of augmentation on the representation space and visualization is not considered \cite{visual_inter}. Li et al. \cite{visual_inter} also provide Explainable CLIP (ECLIP) as an extension to CLIP that uses masked max pooling technique to avoid semantic shift and provide extensive experimentation for their new methodology but still lack the shift change analysis for single input multiple augmentations.

Madasu et al. \cite{Madasu2024QuantifyingAE} propose CLIP-InterpreT tool capable of analyzing property based neighbor search, per-attention head topic segmentation, contrastive segmentation, per-head nearest neighbors of image and per-heat nearest neighbors of text. Similar to Li et al \cite{visual_inter}, CLIP-InterpreT provides new insights into interpretability of the model but doesn't dive deeper into multiple robust quantitative statistical analyses or provide quantitative result for multiple augmentations of the same image.

Kalibhat et al. \cite{pmlr-v202-kalibhat23a} propose FALCON, a framework to explain features of image representation for CLIP. Similar to \cite{intel}, the authors have provided insights into explainability of CLIP and discuss debugging failures in downstream tasks. Similar to the previous research, the authors here too have not considered the impact of augmentation on single image representation and its shift under those constraints.

We position our paper in this gap where robustness on mechanical interpretability for VLMs like CLIP is missing key quantitative analysis, and entirely missing the study of representation shift on image transformations. 

\section{Method}\label{sec:methodology}

\begin{figure*}[!t]
    \centering
    
    \begin{subfigure}{\textwidth}
        \centering
        \includegraphics[width=\textwidth]{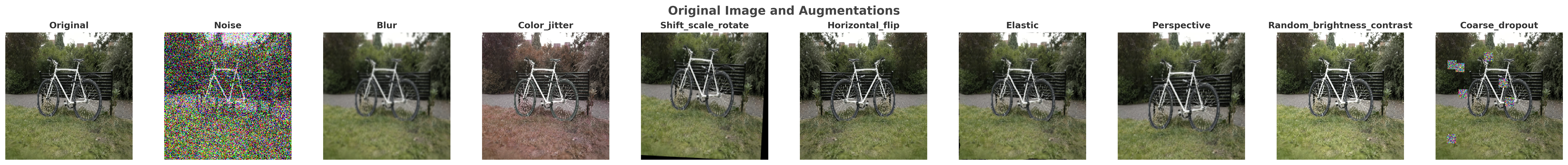}
        \caption{Image augmentation sample}
        \label{fig:augmentation_sample}
    \end{subfigure}
    
    
    \begin{subfigure}{0.33\textwidth}
        \centering
        \includegraphics[width=\textwidth, height=4.6cm]{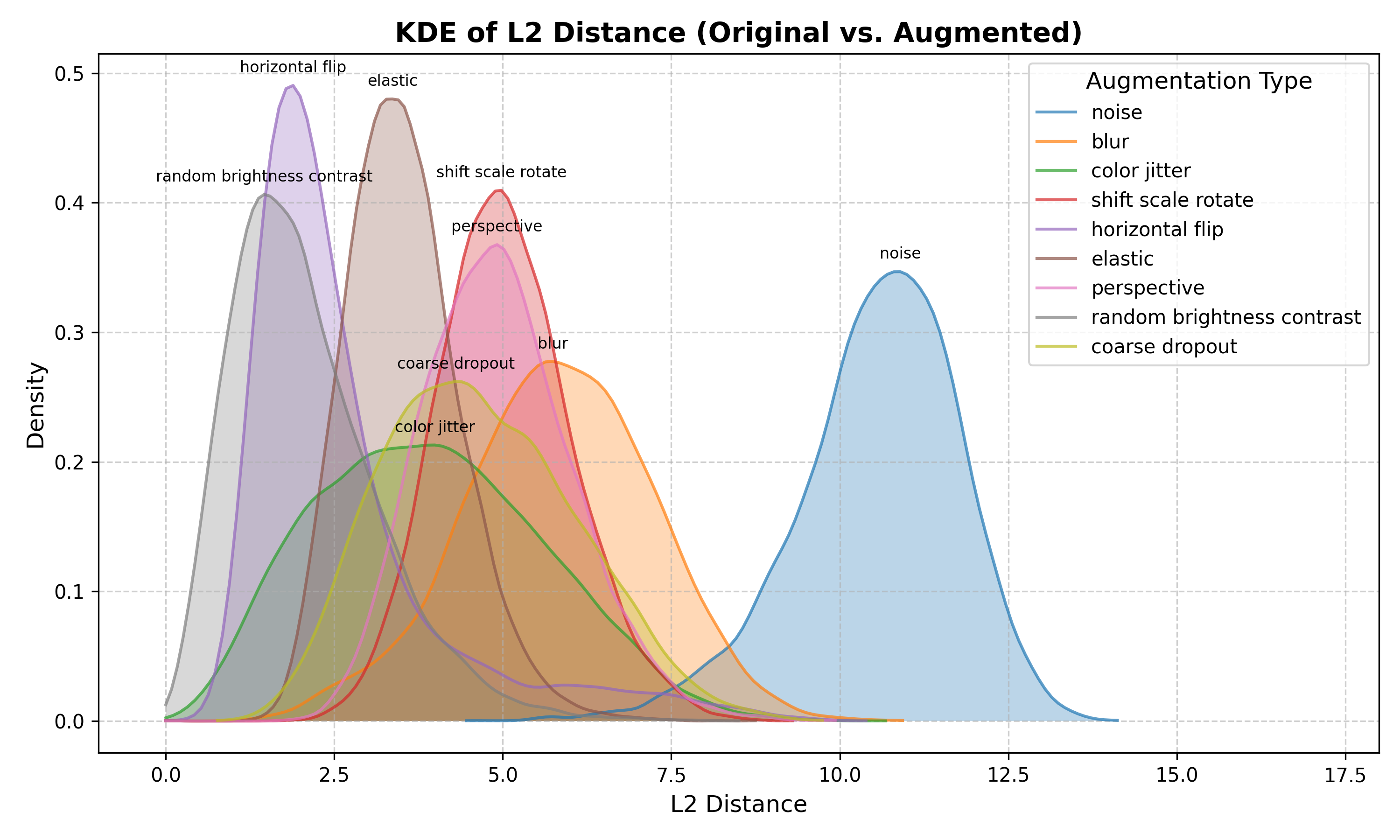}
        \caption{Kernel Density Estimation for L2 Distance}
        \label{fig:kde}
    \end{subfigure}
    \hfill
    \begin{subfigure}{0.33\textwidth}
        \centering
        \includegraphics[width=\textwidth, height=4.6cm]{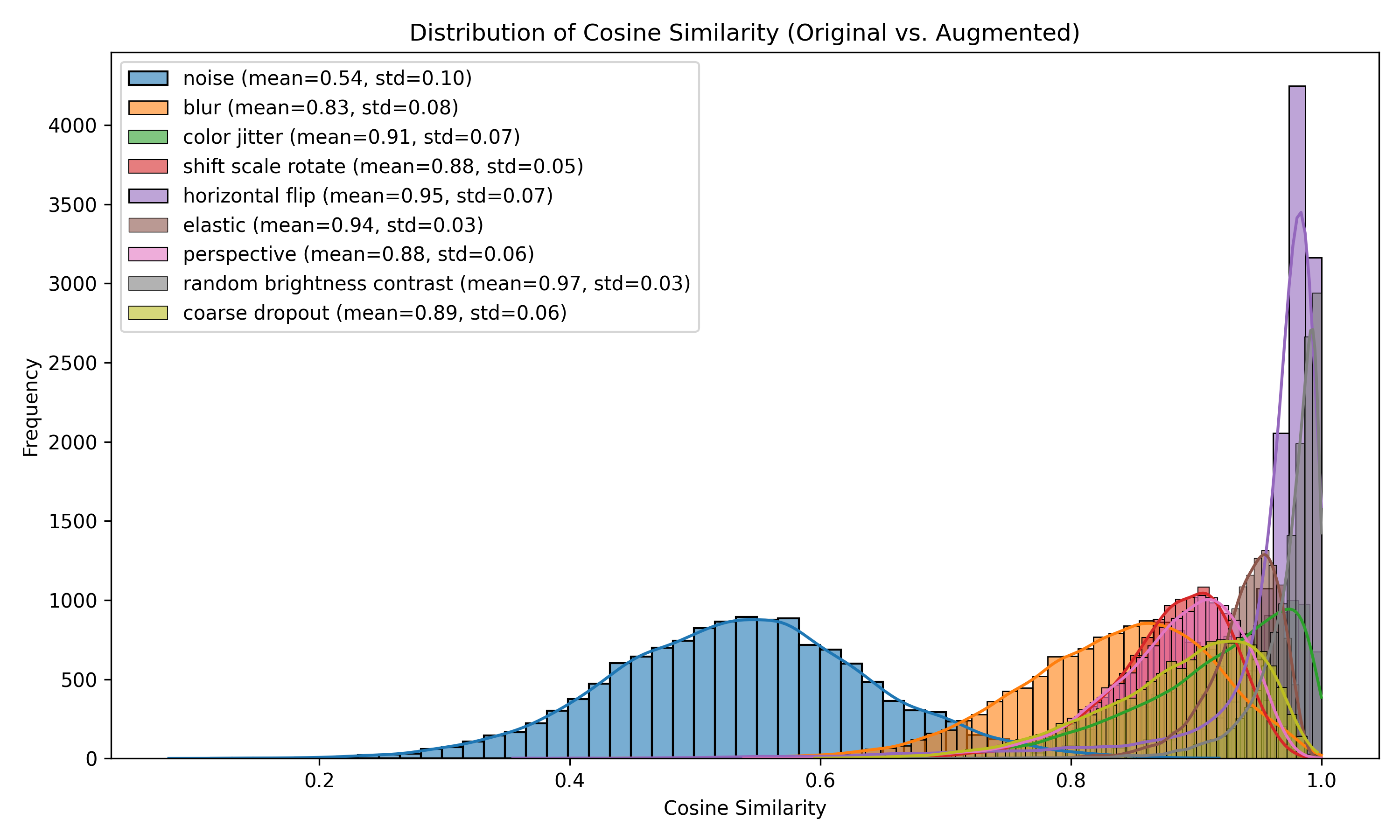}
        \caption{Cosine Similarity Distribution}
        \label{fig:similarity}
    \end{subfigure}
    \begin{subfigure}{0.33\textwidth}
        \centering
        \includegraphics[width=\textwidth, height=4.6cm]{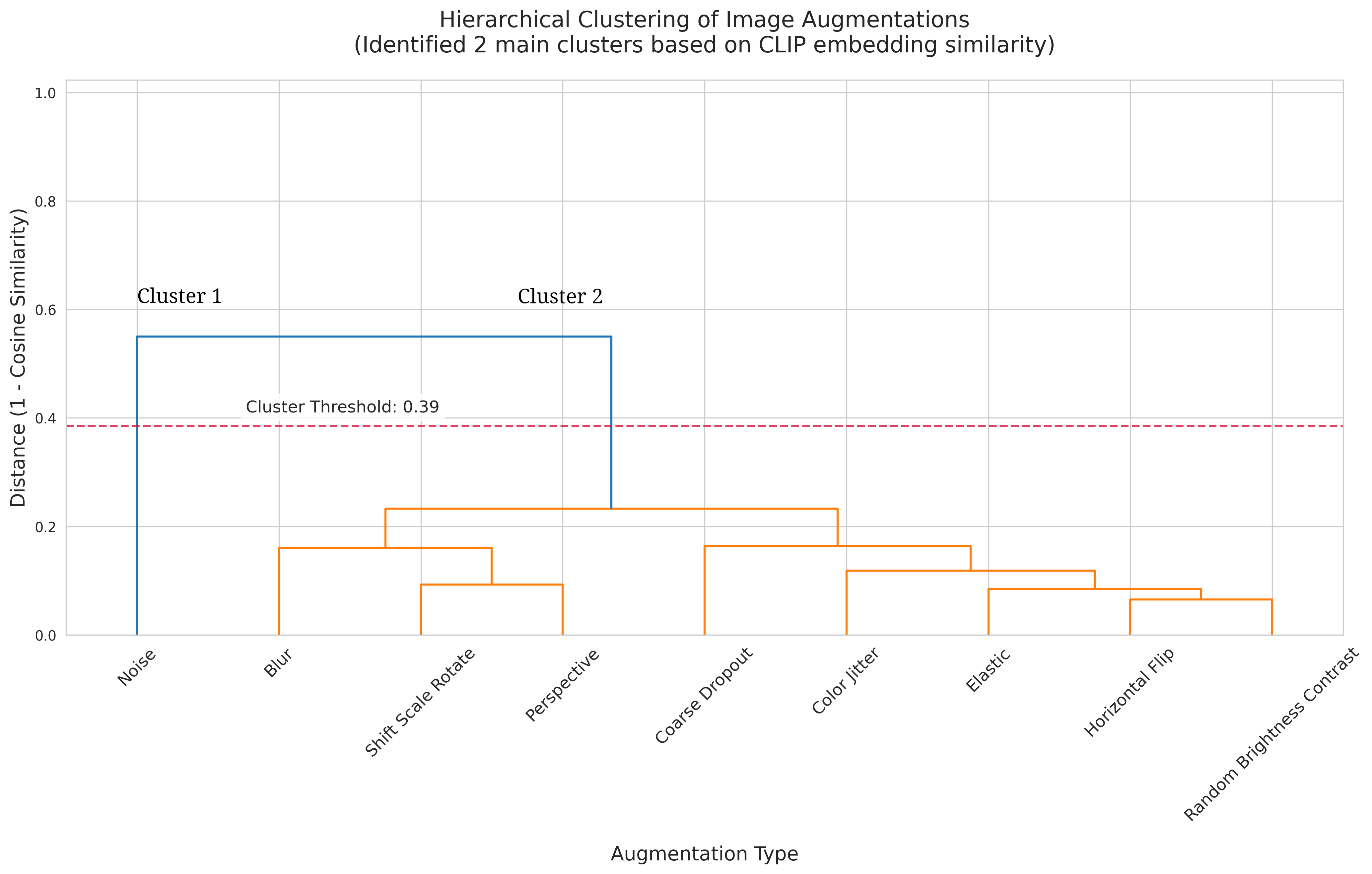}
        \caption{Hirearchal Dendrogram Clustering}
        \label{fig:dendrogram}
    \end{subfigure}

    \caption{Overall Analysis of Proposed Methodology.}
    \label{fig:main_results}
\end{figure*}

\subsection{Dataset}
We use a subset of 13,312 images from the validation set of the Conceptual Captions dataset \cite{conceptual_captions} to implement the methodology presented in \cref{fig:methodology}. We further subsample 2,000 images for the additional metrics presented on \cref{fig:methodology}. Conceptual Captions was chosen as the ideal dataset for the task as the dataset contains random images from the internet and is often used for image captioning with VLM models like BLIP \cite{blip}.

\subsection{Augmentations}
The 9 augmentations presented in \cref{fig:methodology} were implemented on albumentations \cite{albumentations}. We used random noise fill for coarse pixel dropouts. The standard deviation range for noise was from 0.44 to 0.88 and the scale limit for perspective transform was from 0.05 to 0.1. Hyperparameters for rest were fixed and are discussed on \cref{app:methodology}.

\subsection{Metrics}
We provide both quantitative statistical analysis and qualitative visualization for the approach in \cref{fig:methodology}. We define and provide the analysis over dendrogram clustering on distance, augmentation analysis over embedding shift, attention shift, patch similarity, edge preservation and detail preservation, Kernel Density Estimation of L2 distance, per-augment distribution of cosine similarity alongside qualitative analysis of the final layer of the attention layer. Discussion of each formulae on \cref{fig:methodology} is on \cref{app:methodology}.

\begin{figure*}[!t]
    \centering
    \includegraphics[width=\linewidth]{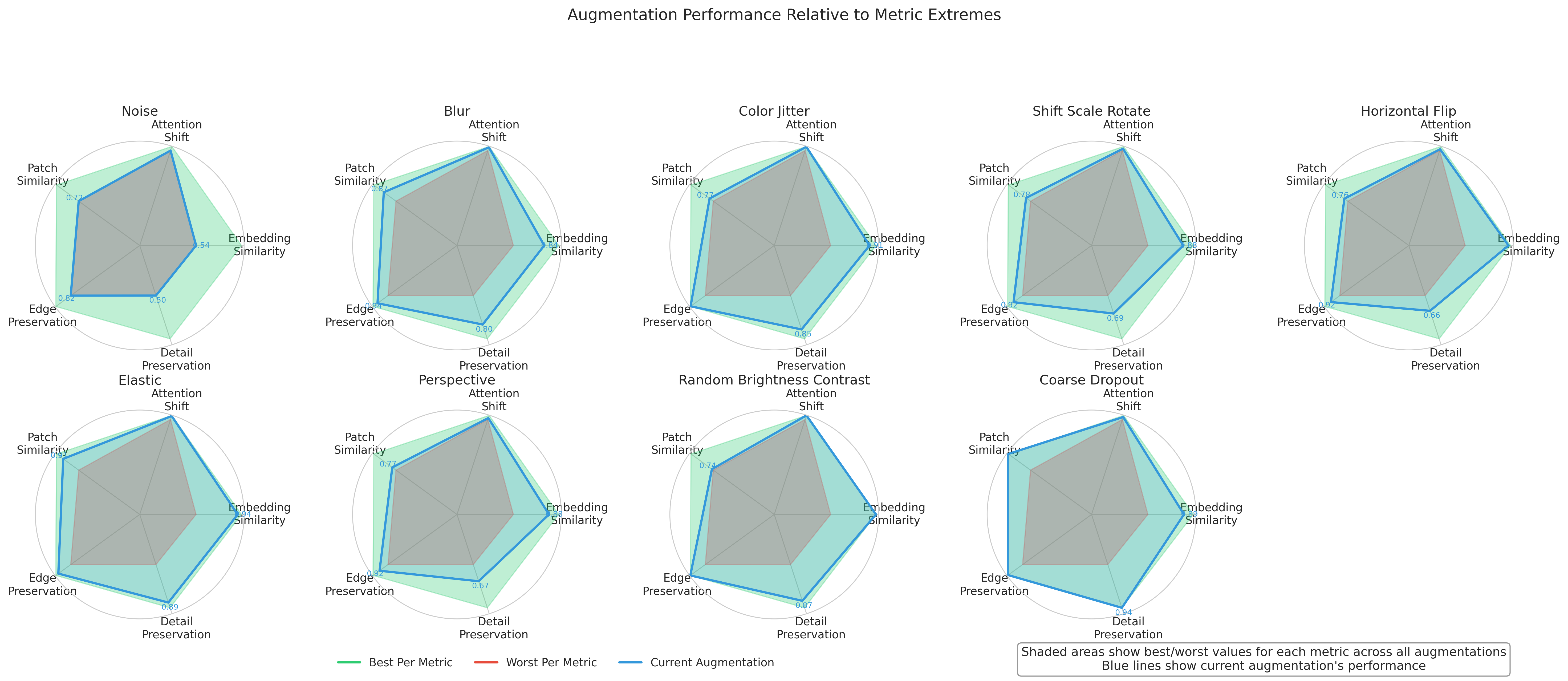}
    \caption{Contextualized radar plot of additional metrics}
    \label{fig:radar}
\end{figure*}

\section{Experiments}\label{sec:experiment_results}

We employ OpenAI's CLIP base patch32 and present our findings. \cref{fig:augmentation_sample} demonstrates our augmentation pipeline and clearly  noise transformation stands apart from the rest of the augmentation, preserving little details that only humans can decipher.

The distribution of attention in \cref{fig:qualitative_results} shows that variability on attention map is mostly diverse on noise augmentation, maps are also highly affected by perspective shifts, blur and shift scale rotation. It also suggests that main object fixation are removed in augmentations like blur, which increases the heatmap for attention, shift scale rotation and perspective transformation, both of which spread the attention focus from the primary subject. These observation on qualitative analysis are further supported by dendrogram clustering in \cref{fig:dendrogram} which used average distance between original embeddings vs augmented embeddings to group them in clusters. Although noise stood apart as it's own cluster we can further notice that even the Cluster 2 has its own subset of two tightly knit together clusters in \cref{fig:dendrogram}, consisting of the blur, and shift scale and perspective rotation.

Moreover, in the KDE and cosine similarity plots also set noise apart from the rest of the augmentation \cref{fig:kde} and \cref{fig:similarity}. Contrastive augmentations, like that of brightness contrast, horizontal flip, elastic or perspective shift have minimal embedding shift both in terms of similarity and distance indicating CLIP's invariability towards color-invariant representations. Whereas color variant augmentations like blur, coarse dropout of pixels and color jitter show higher degree of variance in both analyses on \cref{fig:kde} and \cref{fig:similarity}.

Apart form the L2 distance and cosine similarity in its own, we contextualize our results in terms of attention shift, patch similarity, edge preservation and detail preservation in each of the augmented images alongside the embedding cosine similarity on the radar plot in \cref{fig:radar} where each augmentation is compared against the best and worst metrics. The only reason horizontal flip doesn't perform as good as the rest of color-invariant augmentation in \cref{fig:radar} is the way patches are compared in detail preservation in \cref{fig:methodology}. Since each patch of image is compared to the other corresponding patch we note asymmetrical images don't perform well on such an evaluation; it would be more useful for augmentations that work on transforming the spatial structure of the image like shift scale rotation, perspective rotation and elastic transformation.

Sorting \cref{fig:radar} as a heatmap based on average performance, we can further see the clearer division of the effects each augmentation have on the representation shift on \cref{fig:density_sorted}. This shows a strong directly proportional correlation between the detail and edge preservation and embedding similarity, except for horizontal shift in terms of detail preservation as discussed previously.
\section{Conclusion and Future Works}\label{sec:conclusion}

Our work systematically analyses CLIP's mechanistic interpretability under 9 different augmentation techniques. Our finding suggest the embedding shift of CLIP's representation for an image is most affected by noise addition, followed by color-variant transformations (blur, coarse dropout and color jitter) and shift scaled rotation and perspective shift while having least impact by contrastive augmentation like brightness contrast, horizontal flip or elastic transformation. This findings provide strong evidence that CLIP's vision encoder doesn't treat all augmentation methods equally, hinting at underlying mechanism on its feature representation. We successfully exploit the structure of CLIP's representation space to provide pathway for further research on interpreting which of these features are either learned or memorized.

\textbf{Limitations and directions for future work:}
Future work on CLIP's mechanistic interpretability could explore more in depth the learned representation on a layer-wise fashion. The cross-model alignment with text could be explored to measure if the representation shift correlate with text keywords involving the type of augmentation being performed in the image. Expansion related to this and previous works by \cite{visual_inter,intel} could result on neighborhood structure in embedding space for image augmentation. Our findings are limited to CLIP; therefore, exploring if the embedding shift also occur in a similar fashion on other VLMs like BLIP\cite{blip}, Kosmos-2 \cite{kosmos2} and Flamingo \cite{flamingo} could offer new perspective on the mechanistic interpretability of multimodal systems. More advanced shift on images, like style change on Stable Diffusion, domain shifts and adversarial data attacks, would also add a dynamic perspective into understanding VLMs behavior towards more complex form of image transformations.

{
    \small
    \bibliographystyle{ieeenat_fullname}
    \bibliography{main}
}

\appendix
\section{Appendix}
We present additional details about our experiment, results and visualizations on the appendix section.
\subsection{Hyperparameters and Metrics Details}
\label{app:methodology}
This section contains the explanation of each variables used on the methodology figure on \cref{fig:methodology}. The custom metrics section contains metrics that are commonly used by multiple algorithms and research works in recent academia.
\subsubsection{SciPy Functions}
Cosine Similarity and L2 distance functions were implemented on numpy but are mentioned in this section as they closely align with SciPy's available implementations. Rest of the metrics like fcluster (the dendrogram) the pdist and squareform were used directly from SciPy without any additional modifications.
\begin{equation}
    C = \texttt{fcluster}(Z, t, \texttt{criterion}='\texttt{distance}')
\end{equation}
where $Z$ is the linkage matrix and $t$ is the distance threshold.

\begin{equation}
    D = \texttt{pdist}(X, \texttt{metric}=m)
\end{equation}
where $X$ is an $n \times m$ matrix and $m$ is the distance metric.

\begin{equation}
    S = \texttt{squareform}(D)
\end{equation}
Converts between condensed and square distance matrices.

\begin{equation}
    \text{sim}(u, v) = \frac{u \cdot v}{\|u\| \|v\|}
\end{equation}
Cosine similarity between vectors $u$ and $v$.

\begin{equation}
    d(u, v) = \sqrt{\sum_{i=1}^n (u_i - v_i)^2}
\end{equation}
L2 distance (Euclidean) between vectors $u$ and $v$.

\subsubsection{Custom Metrics}
Our inspiration for these metrics were both derived form previous works \cite{intel,visual_inter,Madasu2024QuantifyingAE} as well as recent industry use of such metrics.
\begin{equation}
    \text{attn}_{\text{sim}} = \frac{1}{1 + \frac{1}{mn}\sum_{i=1}^m \sum_{j=1}^n (A_{ij} - A'_{ij})^2}
\end{equation}
where $A$ and $A'$ are original and augmented attention maps of size $m \times n$.

\begin{equation}
    \text{patch}_{\text{sim}} = \frac{1}{16} \sum_{k=1}^{16} \frac{1}{1 + \frac{\text{MSE}_k}{255^2}}
\end{equation}
where for each patch:
\begin{equation}
    \text{MSE}_k = \frac{1}{whc} \sum_{i=1}^w \sum_{j=1}^h \sum_{c=1}^3 (P_{ijk} - P'_{ijk})^2
\end{equation}

\begin{figure*}[!t]
    \centering
    \includegraphics[width=.75\textwidth]{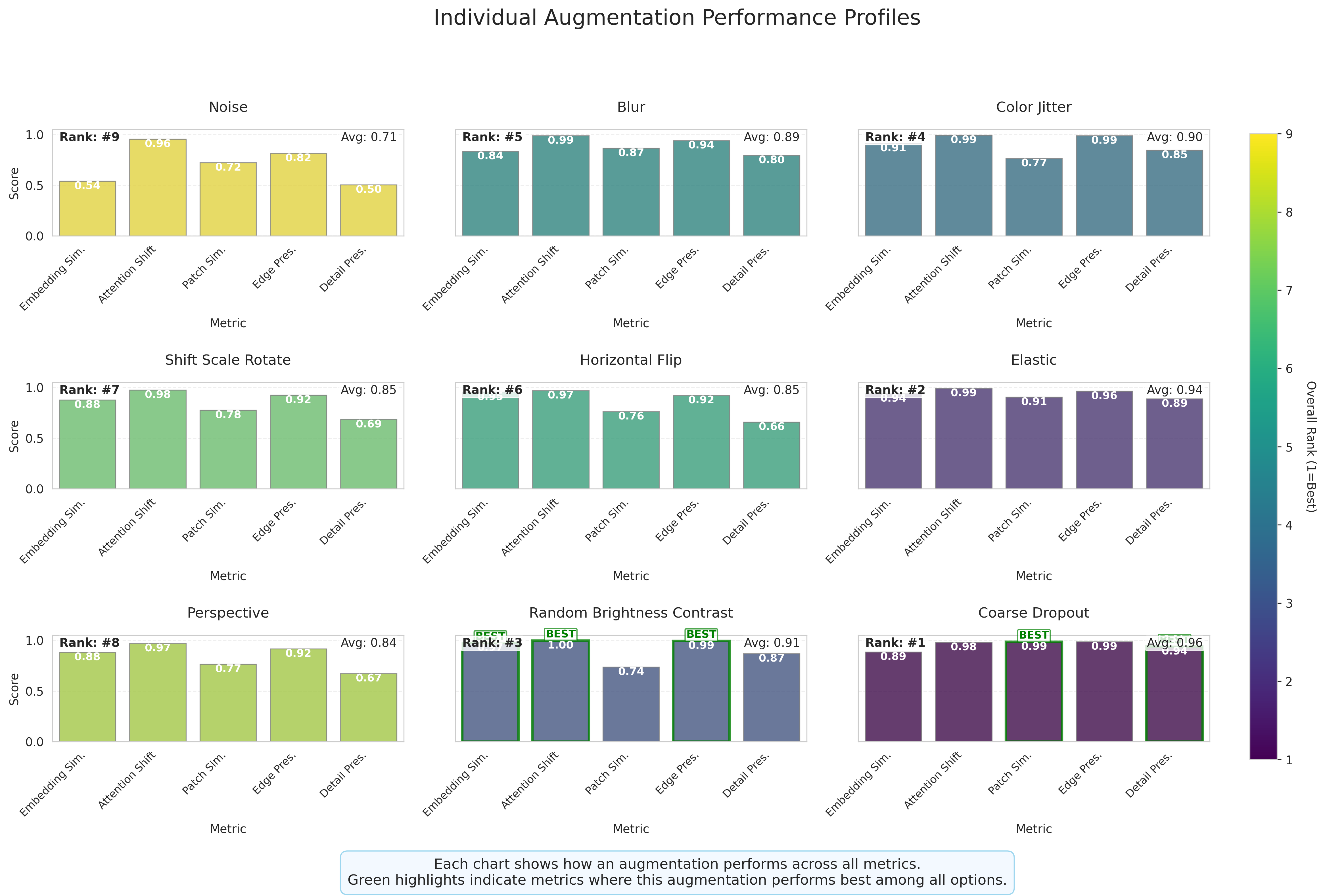}
    \caption{Ranked bar plot for each augmentation profile on performance metrics}
    \label{fig:augmentation_profile}
\end{figure*}

\begin{equation}
    \text{edge}_{\text{sim}} = 1 - \frac{1}{mn} \sum_{i=1}^m \sum_{j=1}^n |E_{ij} - E'_{ij}|
\end{equation}
where edge maps are computed as:
\begin{align}
    E &= \frac{|G_x(I)| + |G_y(I)|}{\max(E)}, \\
    E' &= \frac{|G_x(I')| + |G_y(I')|}{\max(E')}
\end{align}

\begin{equation}
    \text{detail}_{\text{sim}} = \frac{1}{N} \sum_{k=1}^N \exp\left(-\left|\log\left(\frac{\sigma(P'_k)}{\sigma(P_k)}\right)\right|\right)
\end{equation}
where:
\begin{itemize}
    \item $P_k$ and $P'_k$ are corresponding patches
    \item $\sigma$ is the standard deviation operator
    \item $N$ is the number of valid patches (excluding uniform ones)
\end{itemize}

\subsubsection{Implementation Notes}
\begin{itemize}
    \item All metrics are averaged over multiple samples; SciPy functions were averaged over all the 13k images whereas custom metrics were averaged over 2,000 unique samples
    \item Image dimensions: $h \times w$ for height and width
    \item Grayscale conversion uses $\text{Gray} = 0.299R + 0.587G + 0.114B$
    \item Gradient operators $G_x$ and $G_y$ are implemented via finite differences
    \item Patch operations use integer division for grid creation
\end{itemize}

\subsubsection{Augmentations Details}
\cref{alg:image_transform} presents our algorithm on the hyperparameters related to augmentation of images. We show the entire logic for our current implementation of the code for the custom dataset as well as the various hyperparameters that were passed on to albumentations \cite{albumentations} to create our unique images.

\begin{algorithm*}[t]
\caption{Image Transformation Dataset Processing}
\label{alg:image_transform}
\begin{algorithmic}[1]
\Procedure{ImageTransformDataset}{$image\_dir$, $transforms\_dict$, $image\_size$}
    \State \textbf{Initialize:}
    \State $\text{self.image\_dir} \gets \text{Path}(image\_dir)$
    \State $\text{self.image\_paths} \gets \text{Collect image paths (**.jpg, **.jpeg, **.png)}$
    \State $\text{Print dataset size: } |\text{self.image\_paths}|$
    
    \State \textbf{Base Transform:}
    \State $\text{self.base\_transform} \gets \text{Resize}(height=image\_size[0], width=image\_size[1])$
    
    \If{$transforms\_dict = \emptyset$}
        \State \textbf{Set default transformations:}
        \State $\text{self.transforms\_dict} \gets \{$
            $\begin{aligned}
                & \text{(1) GaussNoise}(std=(0.44,0.88), p=1.0), \\
                & \text{(2) GaussianBlur}(kernel=(3,7), p=1.0), \\
                & \text{(3) ColorJitter}(brightness/contrast/saturation/hue=0.2, p=1.0), \\
                & \text{(4) ShiftScaleRotate}(shift=0.0625, scale=0.1, rotate=15^\circ, p=1.0), \\
                & \text{(5) HorizontalFlip}(p=1.0), \\
                & \text{(6) ElasticTransform}(\alpha=30, \sigma=60, p=1.0), \\
                & \text{(7) Perspective}(scale=(0.05,0.1), p=1.0), \\
                & \text{(8) RandomBrightnessContrast}(limit=0.2, p=1.0), \\
                & \text{(9) CoarseDropout}(num\_holes=6-8, size=16\times16, fill=\text{random}, p=1.0)
            \end{aligned}$
        \}
    \Else
        \State $\text{self.transforms\_dict} \gets transforms\_dict$
    \EndIf
\EndProcedure

\Function{GetItem}{$idx$}
    \State $image\_path \gets \text{self.image\_paths}[idx]$
    \State $image \gets \text{ReadRGB}(image\_path)$
    \State $original \gets \text{ApplyTransform}(image, \text{self.base\_transform})$
    
    \State Initialize result dictionary:
    \State $\textit{result} \gets \{
        \begin{aligned}
            & \text{"image\_path"} \colon image\_path, \\
            & \text{"original"} \colon original
        \end{aligned}
    \}$
    
    \For{\textbf{each} $(name, transform)$ \textbf{in} $\text{self.transforms\_dict}$}
            
        \State $transformed \gets \text{ApplyTransform}(original, transform)$
    \EndFor
    \State \Return $\textit{result}$
\EndFunction
\end{algorithmic}
\end{algorithm*}

\begin{figure*}
    \centering
    \includegraphics[width=.48\linewidth]{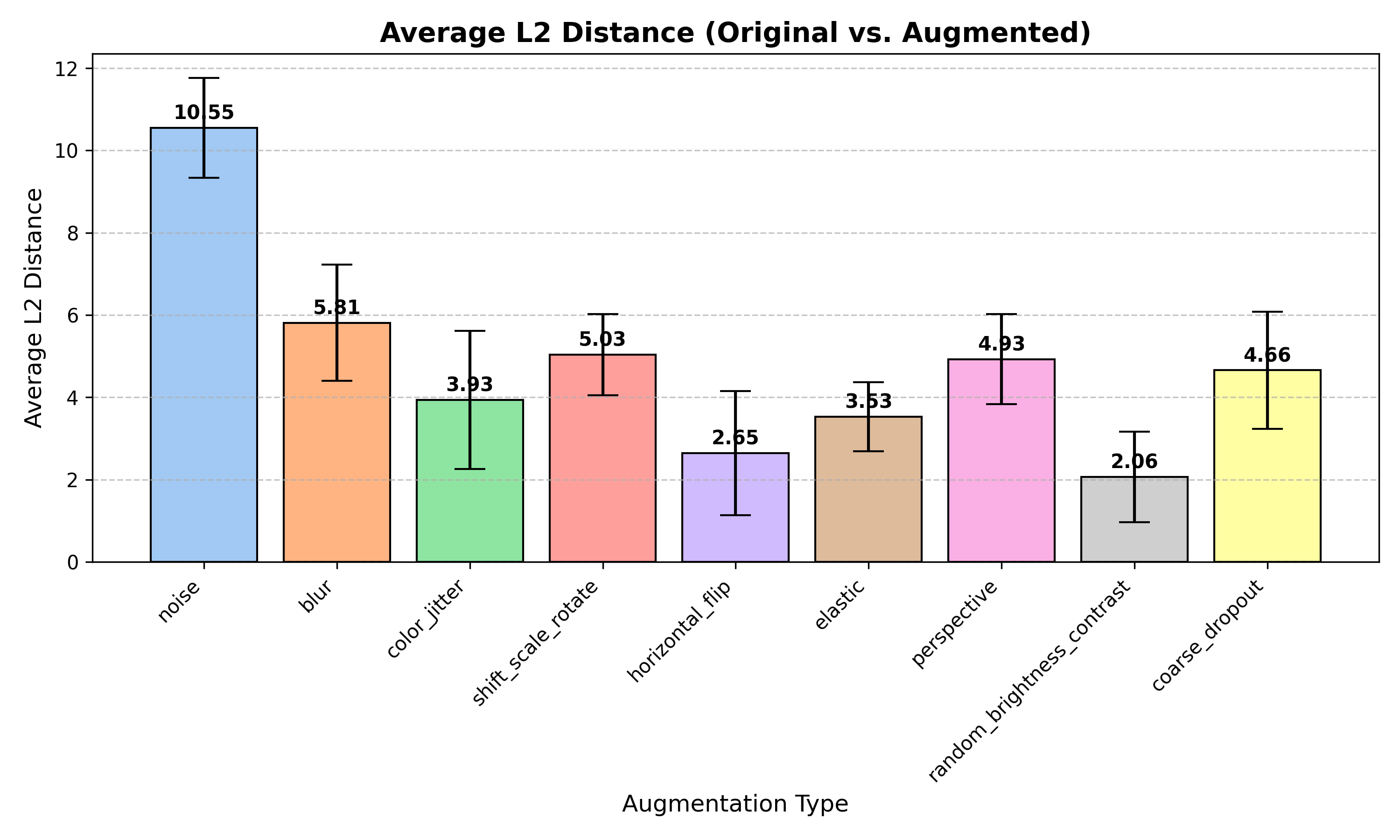}
    \caption{Average L2 distance bar plot for each metric}
    \label{fig:average-distance}
\end{figure*}

\begin{figure*}[t]
    \centering

    \vspace{0.5cm} 
    
    \begin{subfigure}[t]{0.47\textwidth}
        \centering
        \includegraphics[width=\textwidth]{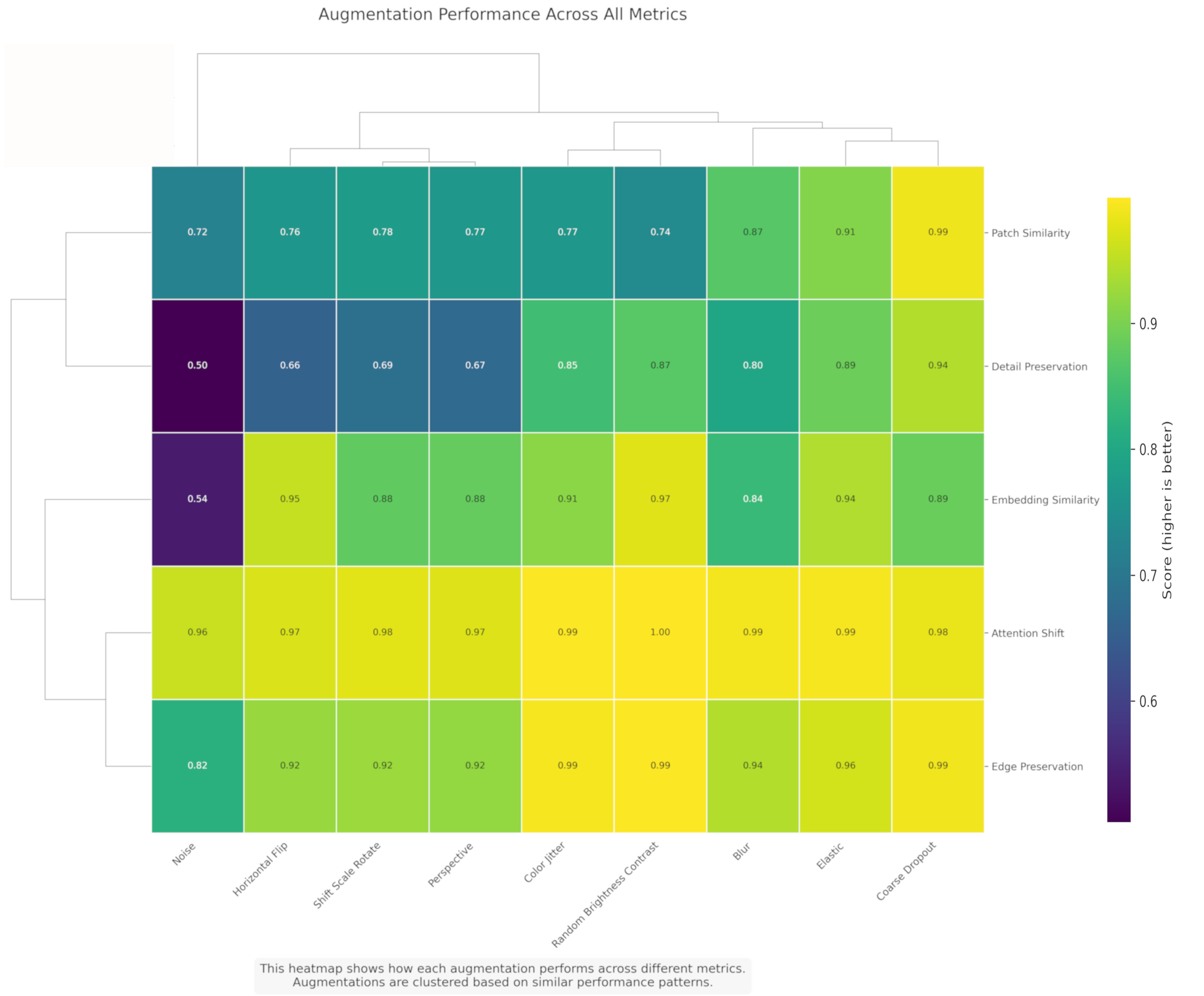}
        \caption{Heatmap with dendrogram clustering over evaluation metrics}
        \label{fig:metics_heatmap}
    \end{subfigure}
    \begin{subfigure}[t]{0.47\textwidth}
        \centering
        \includegraphics[width=\textwidth]{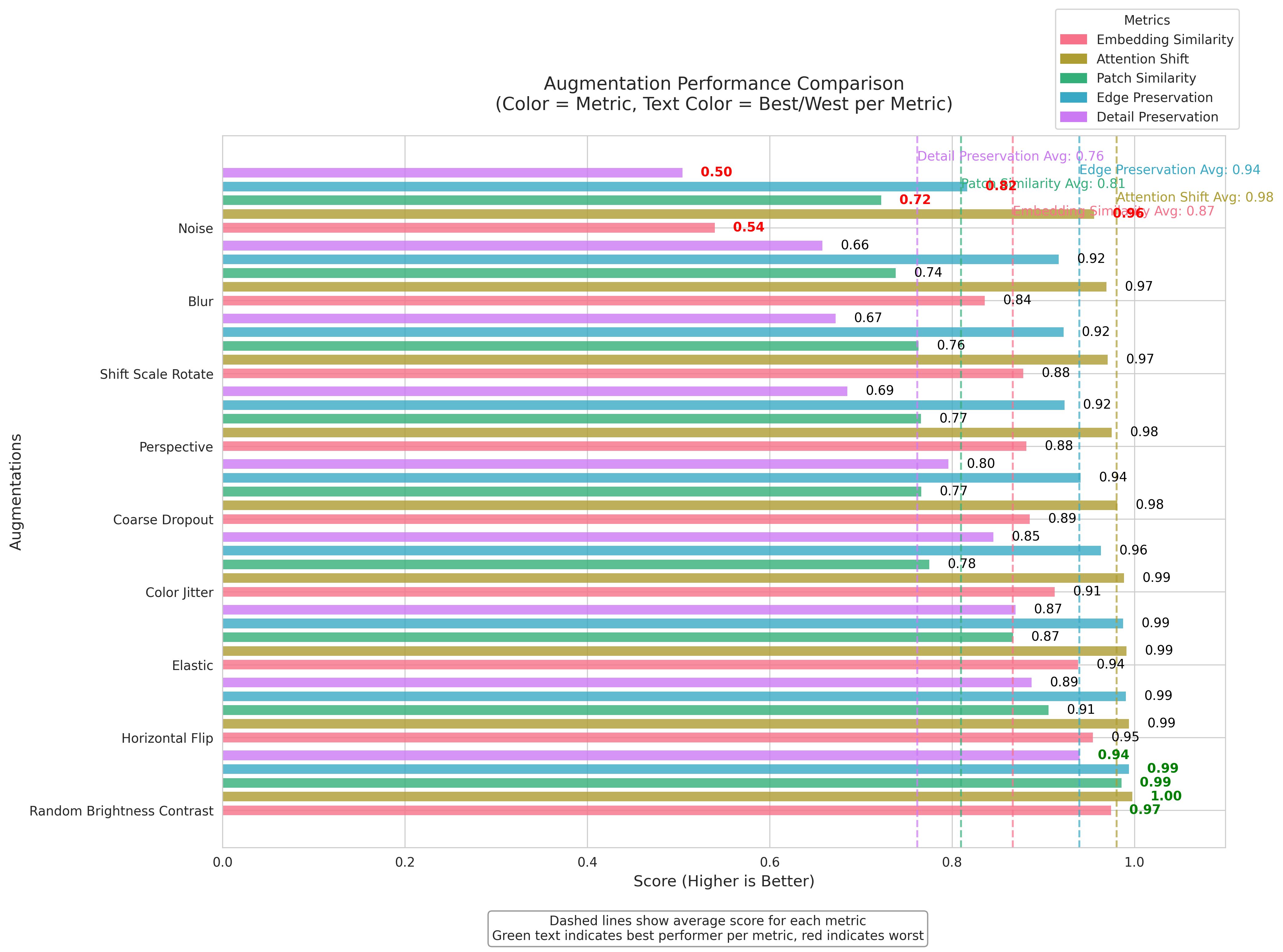}
        \caption{Average augmentation performance comparison on custom metrics}
        \label{fig:augmentation_comparison}
    \end{subfigure}
    
    \caption{Additional quantitative analysis results}
    \label{fig:additional_results}
\end{figure*}

\subsection{Additional Results}

\begin{figure*}[!t]
    \centering
    \includegraphics[width=.7\linewidth]{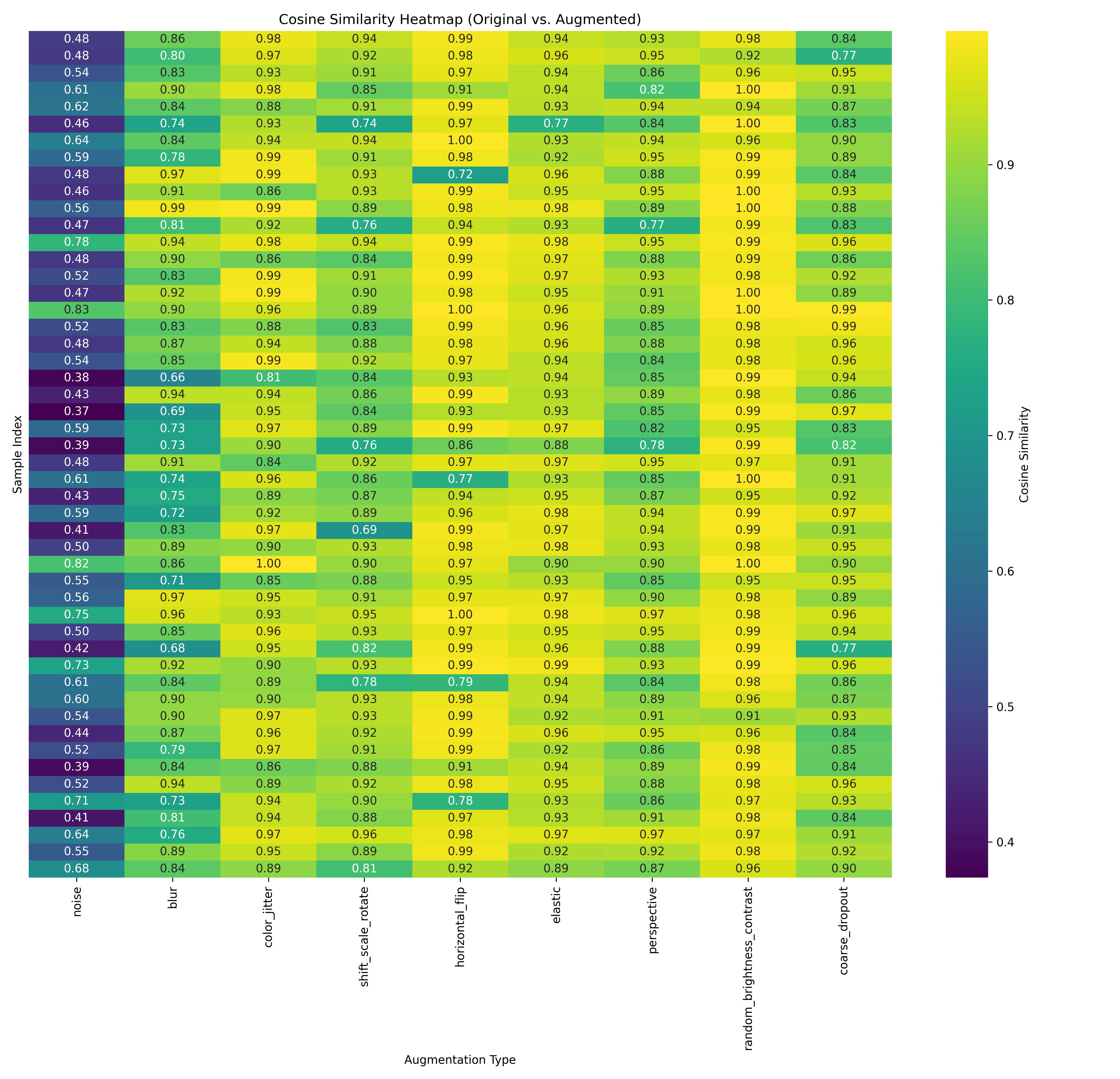}
    \caption{Heatmap of 50 random sample's cosine similarity towards the embedding of original image}
    \label{fig:50-heatmap}
\end{figure*}

We present a new perspective to the results using different graphs for the quantitative results observed in \cref{sec:experiment_results} and provide more in-depth examples of qualitative results in this section in \cref{fig:average-distance,fig:additional_results,fig:50-heatmap}.
In \cref{fig:average-distance}, we show the average L2 distance of each augmentation's embeddings against the original embeddings. This is a further intuitive explanation of the KDE plot in \cref{fig:kde}. \cref{fig:augmentation_profile} shows a rank fashion bar plot ranking each of the augmentation based on average performance across all metrics. It provides more visual intuition towards the results observed in \cref{fig:similarity}.

\cref{fig:metics_heatmap} show a combined intuition towards the dendrogram clustering in \cref{fig:dendrogram} combined together with \cref{fig:density_sorted}. We also took 50 random samples and evaluated the cosine similarity of each sample's augmented representation with the original representation to check for metrics consistency and report it on \cref{fig:50-heatmap}. An unsorted version of \cref{fig:augmentation_profile} that instead highlights the overall average of each metrics is presented on \cref{fig:augmentation_comparison}.

Following pages contain some of the qualitative analysis we have conducted that are an extension of the abstract and visualization for a comprehensive review of the paper \cref{fig:qualitative_results} and \cref{fig:augmentation_sample}.

%
%

\begin{figure*}
    \centering
    \includegraphics[width=.95\textwidth]{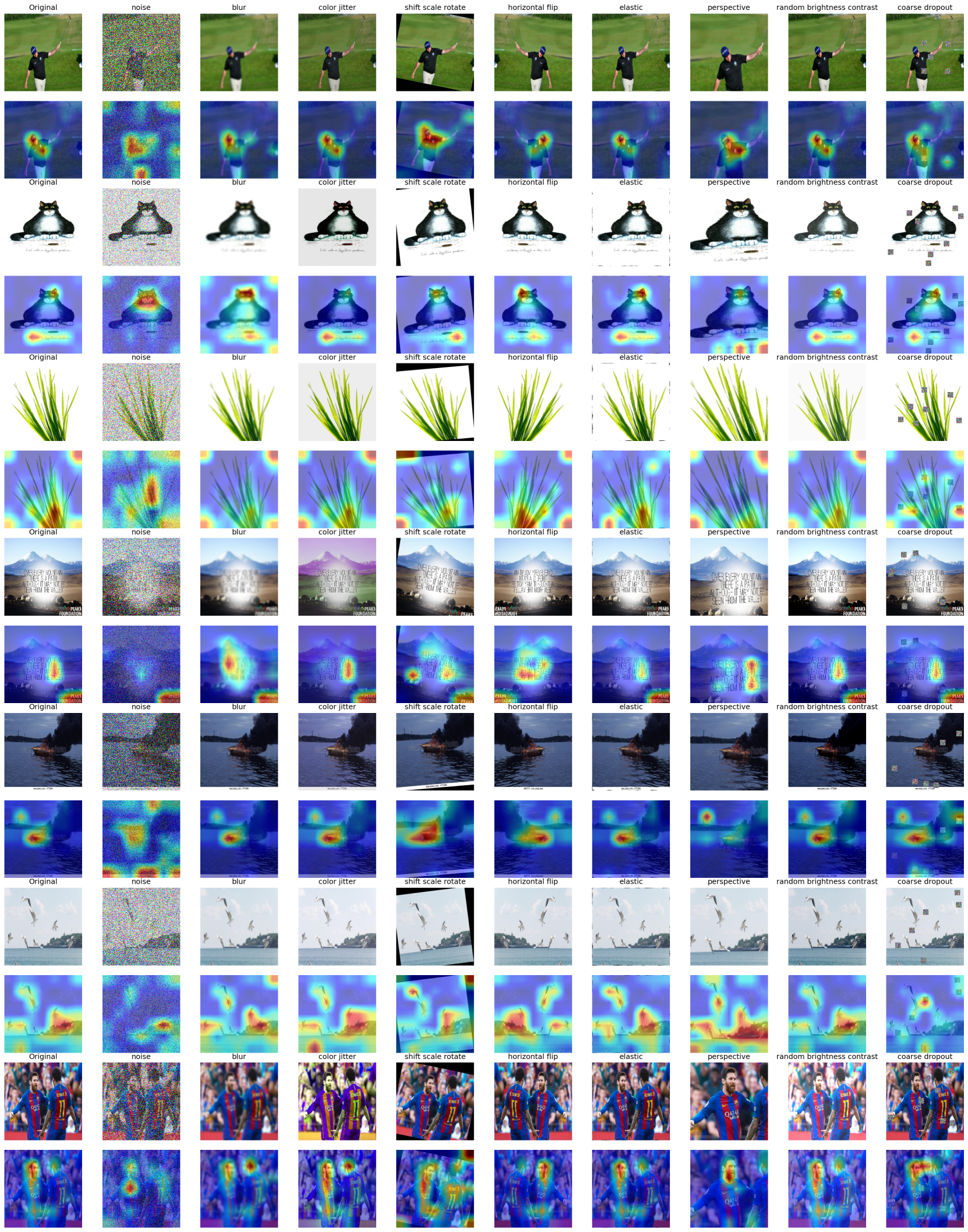}
    \caption{Sample qualitative analysis of attention map for each augmentation types}
\end{figure*}

\begin{figure*}
    \centering
    \includegraphics[width=.95\textwidth]{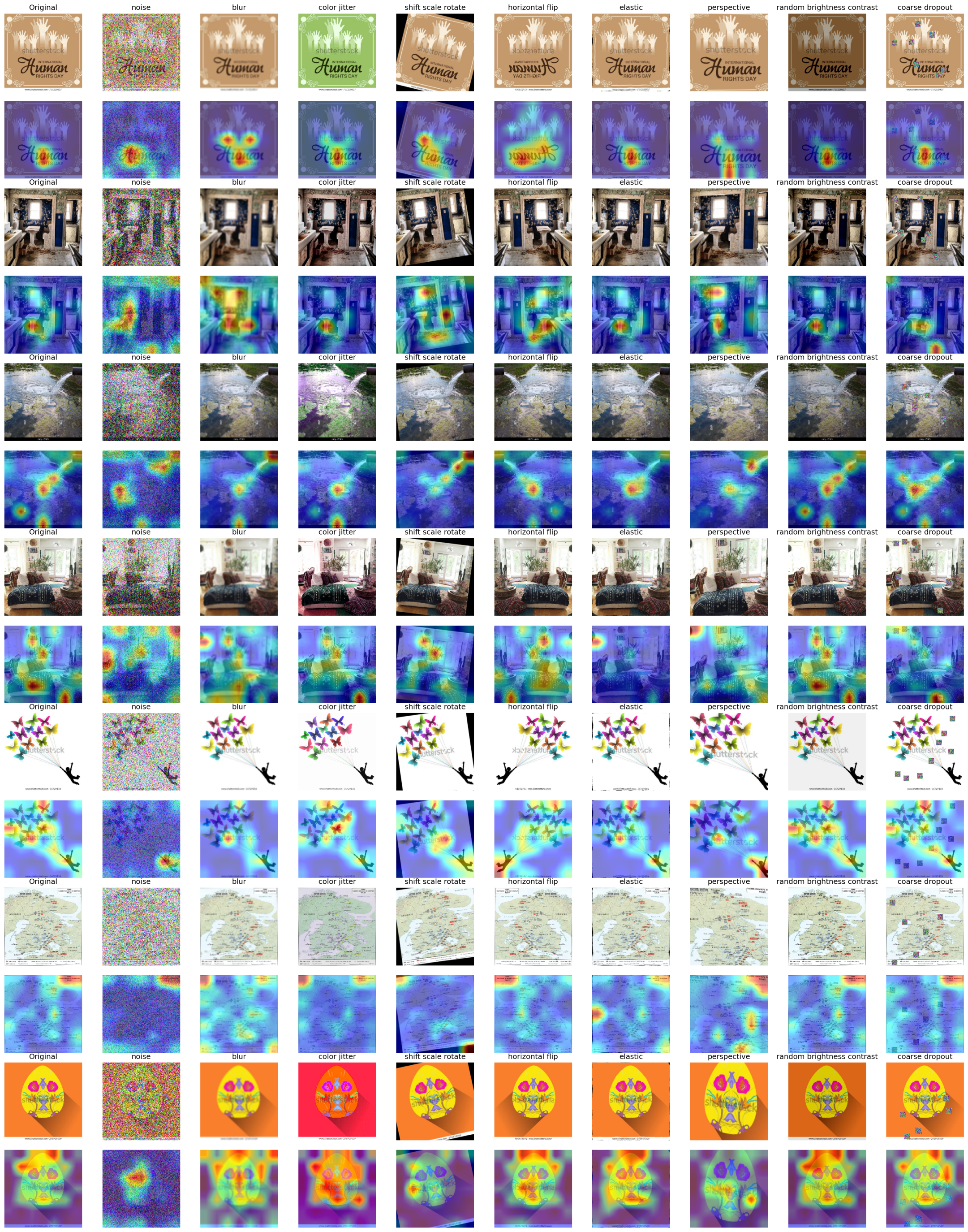}
    \caption{Sample qualitative analysis of attention map for each augmentation types}
\end{figure*}

\end{document}